%% file: main.tex
\documentclass{article}
\usepackage{spconf,amsmath,graphicx}
\usepackage[colorlinks]{hyperref}

\usepackage{tabu}

\title{Exploration of Visual Features and their weighted-additive fusion for Video Captioning}
\name{\begin{tabular}{c}Praveen S V, Akhilesh Bharadwaj,  Harsh Raj, Janhavi Dadhania,  \\
Ganesh Samarth C.A, Nikhil Pareek, S R M Prasanna\end{tabular}}
\address{Indian Institute of Technology Dharwad, UltraInstinct AI}
\begin{document}
%

\begin{minipage}[b]{0.9\textwidth}
\Large \textbf{Preprint Notice:} \\
\\
\large
© 2021 IEEE. Personal use of this material is permitted. Permission from IEEE must be obtained for all other uses, in any current or future media, including reprinting/republishing this material for advertising or promotional purposes, creating new collective works, for resale or redistribution to servers or lists, or reuse of any copyrighted component of this work in other works.
\end{minipage}

\input{tex/1.abstract.tex}
\input{tex/2.introduction.tex}
\input{tex/3.related_works.tex}
\input{tex/4.proposed_approach.tex}
\input{tex/5.experimental_results.tex}
\input{tex/6.conclusion.tex}

\small{
\bibliographystyle{IEEEbib}
\bibliography{ref/refs}
}

\end{document}

%% file: tex/1.abstract.tex
\maketitle
\begin{abstract}
    Video captioning is a popular task that challenges models to describe events in videos using natural language. In this work, we investigate the ability of various visual feature representations derived from state-of-the-art convolutional neural networks to capture high-level semantic context. We introduce the Weighted Additive Fusion Transformer with Memory Augmented Encoders (WAFTM), a captioning model that incorporates memory in a transformer encoder and uses a novel method, to fuse features, that ensures due importance is given to more significant representations. We illustrate a gain in performance realized by applying Word-Piece Tokenization and a popular REINFORCE algorithm. Finally, we benchmark our model on two datasets and obtain a CIDEr of 92.4 on MSVD and a METEOR of 0.091 on the ActivityNet Captions Dataset.
\end{abstract}

\begin{keywords}
video captioning, visual features, waftm, WordPiece, REINFORCE
\end{keywords}

%% file: tex/2.introduction.tex
\section{Introduction}
\label{sec:introduction}

Solving the challenge of video captioning is rewarding due to its tremendous applications in automatic video summarization, innovative technology for the visually-disabled and video-retrieval systems. It is considered an intricate task in the domain of computer vision due to the difficulty involved in mapping visual modalities to human interpretable language. Many approaches \cite{wang2018reconstruction, aafaq2019spatio} tackle this problem by using recurrent neural networks (RNNs) to generate captions from input visual representations. These representations are often derived from 2D or 3D Convolutional Neural Networks (CNNs), pre-trained on various tasks such as image classification or action recognition with the intent of providing enhanced visual understanding.  However, each of these networks captures different kinds of visual context such as motion, appearance and object-level features. While individual representations from different networks can only partially reason the semantics, their aggregation is often more useful in providing a holistic understanding of a video. 




The introduction of transformers \cite{vaswani2017attention}, that solely rely on the attention mechanism, have demonstrated superior and efficient performance in sequential tasks. Some variants \cite{chen2018tvt} of the transformer have achieved competent performance for captioning videos, although, like other sequential models, they too are susceptible to the long-tailed problem commonly occurring in natural language generation tasks. In this problem, a large number of significant words are not adequately learnt by the model owing to their low frequency of occurrence in the corpus. Frequency-based weights \cite{dong2019not} given to each word alleviates the problem to an extent. Other approaches have explored using External Language Models (ELM) like BERT \cite{devlin2018bert} to counter the issue. 

Our main contributions in this work are - 1) An exhaustive feature exploration spanning different state-of-the-art pre-trained models to determine the most suitable representations for video captioning. 2) A weighted-additive fusion technique during encoder-decoder cross attention with the ability to aggregate multiple representations while ensuring each one is given its due importance. 3) Improvisation of captioning model performance by replacing a word-frequency based tokenizer with WordPiece tokenization and training using a variant of the REINFORCE algorithm. 

%% file: tex/3.related_works.tex
\section{Related Works}
\label{sec:related-works}

Current research trends in solving the task of captioning videos adopt a two staged approach : extracting feature representations and employing sequence-to-sequence based architectures to further generate captions. Several works \cite{wang2018reconstruction, chen2018tvt, pei2019memory} utilise 2D CNNs such as Inception-v4 \cite{szegedy2016inception}, NASNet \cite{zoph2018learning}, VGG \cite{simonyan2014very}, ResNet \cite{he2016deep} and its variants to extract features from sampled video frames. Few works have also explored 3D CNNs like Convolutional 3D (C3D) \cite{tran2015learning} and Inflated 3D ConvNets (I3D) \cite{carreira2017quo} to capture motion features from videos. Fusing representations to enhance semantic context can prove to be challenging and a common approach for aggregation is through concatenation \cite {iashin2020multi}. Work in \cite{chen2018tvt, iashin2020better} explores fusion mechanisms based on attention to exploit information from different modalities. 

RNNs have been the preferred architectural choice for most sentence generation tasks. For example, RecNet \cite{wang2018reconstruction} utilizes LSTMs to generate sentences from encoded video semantic features and trains the model further by reconstructing video features. More recent works demonstrate the efficiency of transformers \cite{chen2018tvt, cornia2020meshed} for captioning tasks. The work in \cite{cornia2020meshed} enhances the transformer by introducing meshed connections between encoder and decoder layers and also introduces persistent memory vectors in the encoder for the task of Image Captioning. We draw inspiration from the above architecture to formulate our weighted-additive fusion mechanism. Finally, to improve captioning performance the authors of \cite{rennie2017self} introduce a self-critical sequence training (SCST) algorithm that uses reinforcement learning and directly optimizes on the CIDEr metric.

%% file: tex/4.proposed_approach.tex
\begin{figure*}[t!]
\centering
\includegraphics[width=0.8\linewidth]{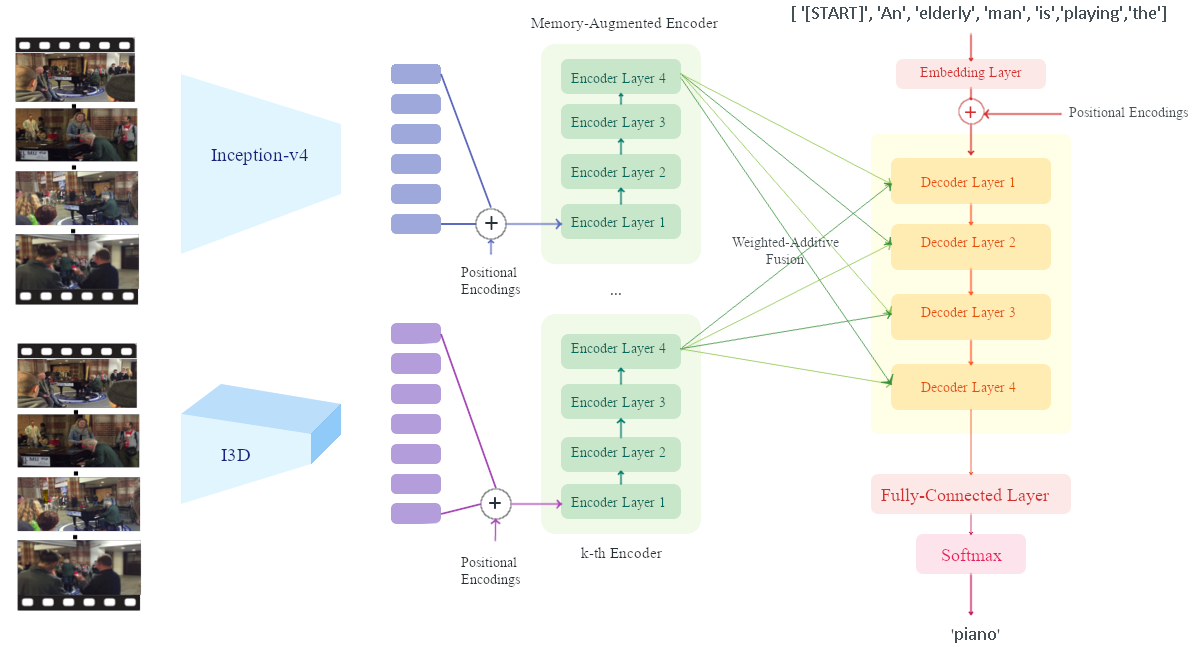}
\vspace{-0.6cm}
\caption{Architecture of the Weighted-Additive Fusion Transformer with Memory-augmented Encoders (WAFTM). First, the input video clips are passed through feature extractors; Inception-v4 and I3D. Next, the visual feature representations, of each modality, are passed through memory-augmented encoders. During encoder-decoder cross attention, weighted-additive fusion is performed at each decoder layer. The output of the final decoder layer is passed through a fully-connected layer followed by a softmax operation to generate a probability distribution over the vocabulary.}
\label{fig:waftm}
\end{figure*}

\section{Proposed Approach}
\label{sec:proposed-approach}
Given, an input sequence of video frames \(v = \{f_t\}\) where \( t \in 1,2,3,...n\) is the temporal index for a video with n frames. The goal is to create a captioning model that generates a sentence, \( S = \{w_j\}, w_j \in V \) where length of sentence varies for each caption of the the video and \(V\) is the vocabulary. 

Our model makes use of multiple memory-augmented encoders \cite{cornia2020meshed} for each input visual modality and performs a parametrized weighted-additive fusion to aggregate the visual representations in the decoder. To obtain the most suitable feature representation of a video, we perform an exhaustive exploration of the following candidate architectures - 

\begin{enumerate}
    \item \textbf{ResNet-152} \cite{he2016deep}: This model is a deep CNN with 152 layers and residual connections. We use it to extract frame-level features.
    
    \item \textbf{Inception V4} \cite{szegedy2016inception}: This network streamlines the original Inception architecture and frame-level features are extracted using an average pooling layer.
    
    \item \textbf{R(2+1)D} \cite{tran2018closer} : This model is a 3D-CNN based on the ResNet-50 architecture that factorizes 3D convolutional filters into separate spatial and temporal components.
    
    \item \textbf{I3D} \cite{carreira2017quo}: The network is a 3D-CNN that inflates the Inception network to learn seamless spatio-temporal feature extractors from videos.
    
    \item \textbf{Faster-RCNN} \cite{ren2016faster}: This model is two-stage object detection network, consisting of a Fast-RCNN with a ResNet-50 backbone and a region proposal network. Features are extracted from the ROI pool layer.
    
\end{enumerate}

\subsection{\textbf{Memory Augmented Encoder}}
The input to the encoder, is a set of input features \( X = \{x_i\}, i = 1,2,..., m \) where $m$ is the input sequence length of the feature representation of the video. Each feature $X$ is passed through a fully connected layer to project it into a D-dimensional space. Here, \( D = 512 \).
\begin{equation}
    X^\prime = W^T_e X + b_e
\end{equation}
The keys and values are obtained by first projecting the embedded features and then extending them with additional memory slots to encode a priori information.
\begin{gather}
    M_{attn} (X^\prime) = Attention(W_q X^\prime, K, V)  \nonumber \\
    K = W_k X^\prime \mathbin \Vert M_k \nonumber \\
    V = W_v X^\prime \mathbin \Vert M_v
\end{gather}
where \(\mathbin \Vert \) indicates concatenation, \(W_e, W_q, W_k, W_v\) are learnable weight matrices, \(M_k, M_v\) are learnable parameters with $m$ rows.

The complete encoding layer consists of memory augmented attention followed by position-wise feedforward, both of which are encapsulated with a residual connection and layer norm operation. We see that, multiple encoder layers can be stacked such that each consumes the output from the previous layer.

\subsection{\textbf{Decoder}} \label{ssec:decoder}

Let $Y \in \mathcal{R}^{l \times 512}$ represent the masked input sequence of $l$ words in the sentence, each of which is projected to a $512$-D vector using a learnable embedding layer. Let $\widehat{X_k}$ represent the output from the last layer of the memory-augmented encoder corresponding to the $k$-th visual feature input modality. Each decoder layer performs cross-attention utilizing the outputs of the $k$ encoders and aggregates them by a parametrized weighted summation.

\begin{equation}
    \Phi(Y, \widehat{X_k}) = Attention(W_q Y, W_k \widehat{X_k}, W_v \widehat{X_k})
\end{equation}

where $\Phi(Y, \widehat{X_k})$ indicates cross-attention output between the $k$-th encoder output and a decoder layer. \\
\\
\noindent \textbf{Weighted Additive Fusion - } To weigh the relative importance of each visual modality and ensure more significant representations are given higher weightage, we learn weights $\alpha_k$ corresponding to the cross-attention output from the $k$-th encoder output and a decoder layer.

\begin{equation}
    \alpha_k = \sigma(W_i [Y \mathbin \Vert \;\Phi(Y, \widehat{X_k})] + b_i)
\end{equation}
where $W_i$ is a learnable weight matrix, $\sigma$ is the sigmoid function, and $b_i$ is the learnable bias vector. The fused feature, $X_{fused}$ is given by, 

\begin{equation}
    X_{fused} = \sum_{k} \alpha_k \; \odot \; \Phi(Y, \widehat{X_k})
\end{equation}

A complete decoder layer consists of memory-augmented attention followed by position-wise feedforward, both of which are encapsulated with a residual connection and a layer norm operation. Multiple decoder layers are stacked, each taking its input as the output from the previous decoder layer. The final decoder layer output, $\widehat{Y}$, is passed through a fully connected layer followed by a softmax operation to obtain a probability distribution over the words of the vocabulary. The complete architecture of WAFTM, comprising of memory-augmented encoders and the decoder, is depicted in Figure \ref{fig:waftm}.

%% file: tex/5.experimental_results.tex
\begin{table*}[ht]
    \centering
    \begin{tabu}{|l|c c c c|}
    \tabucline[1pt]{-}
       \hspace{1.7cm} \textbf{Features}  & \textbf{B@4} & \textbf{R} & \textbf{M} & \textbf{C} \\ 
    \tabucline[1.2pt]{-}
      ResNet-152 & 0.473 & 0.677 & 0.329 & 77.15 \\ 
       Inception-v4 & 0.499 & 0.679 & 0.327 & 81.96  \\
    \hline
       I3D & 0.437 & 0.678 & 0.314 & 69.24 \\ 
       R(2+1)D & 0.504 & 0.700 & 0.336 & 82.41 \\ 
    \hline
       Faster-RCNN & 0.247 & 0.525 & 0.209 & 60.76 \\ 
    \hline
    
       I3D + Faster-RCNN & 0.453 & 0.675 & 0.318 & 69.62 \\ 
       I3D + ResNet-152 & 0.478 & 0.699 & 0.337 & 81.86 \\ 
    \hline
        Inception-v4 + Faster-RCNN & 0.492 & 0.688 & 0.334 & 83.36 \\ 
        Inception-v4 + R(2+1)D & 0.510 & 0.714 & \textbf{0.345} & 88.88 \\ 
        
        Inception-v4 + I3D & 0.501 & \textbf{0.715} & 0.337 & 89.58 \\ 
    \hline
    Inception-v4 + I3D + Faster-RCNN & 0.465 & 0.678 & 0.325 & 81.14 \\
    Inception-v4 + R(2+1)D + Faster-RCNN & \textbf{0.511} & \textbf{0.715} & 0.344 & \textbf{90.93} \\
    \tabucline[1pt]{-}
    \end{tabu}
    \vspace{0.1cm}
    \caption{Comparison of captioning performance using different feature extractors benchmarked using WAFTM + WordPiece on the MSVD dataset.}
    \label{tab:comparing_features}
\end{table*}
\begin{table}[!ht]
    \centering
    \begin{tabu}{|l| c c c c|}
    \tabucline[1pt]{-}
       \hspace{0.85cm}\textbf{Architecture} & \textbf{B@4} & \textbf{R} & \textbf{M} & \textbf{C} \\ \tabucline[1pt]{-}
    \footnotesize RecNet \cite{wang2018reconstruction} & \small 0.523 & \small 0.698 & \small 0.341 & \small 80.3 \\ \tabucline[0.1pt]{-}
       \footnotesize TVT \cite{chen2018tvt} & \small 0.532 & \small 0.720 & \small 0.352 & \small 86.76 \\ \tabucline[0.1pt]{-}
        \footnotesize MARN \cite{pei2019memory} & \small 0.486 & \small 0.719 & \small 0.351 & \small 92.2 \\ \tabucline[0.1pt]{-}
        \tabucline[1pt]{-}
       \footnotesize WAFTM & \small 0.502 & \small 0.708 & \small 0.330 & \small 85.57 \\
       \tabucline[0.1pt]{-}
       \footnotesize WAFTM + WordPiece & \small 0.511 & \small 0.715 & \small 0.344 & \small 90.93 \\ 
       \tabucline[0.1pt]{-}
       \footnotesize WAFTM + WordPiece + SCST & \small 0.504 & \small 0.708 & \small 0.342 & \small \textbf{92.43} \\ \tabucline[1pt]{-}
    \end{tabu}
    \vspace{0.1cm}
    \caption{Results on the official splits of MSVD Dataset.}
    \label{tab:msvd}
\end{table}
\begin{table}[!ht]
    \centering
    \begin{tabu}{|l|c|}
    \tabucline[1pt]{-}
       \hspace{0.8cm} \textbf{Architecture} & \textbf{M} \\ \tabucline[1pt]{-}
       \footnotesize Krishna et. al \cite{krishna2017dense} & \small 0.0888 \\ \hline
       \footnotesize Iashin et. al \cite{iashin2020better} & \small 0.0956 \\ \hline
       \footnotesize WAFTM + WordPiece + SCST & \small 0.0906 \\ \tabucline[1pt]{-}
    \end{tabu}
    \vspace{0.1cm}
    \caption{Results on ActivityNet Captions Dataset using ground-truth proposals and where each model is trained from scratch.}
    \label{tab:activitynet}
\end{table}
\section{Experimental Results}
\label{sec:experimental-results}
We use four popular metrics to evaluate the performance of our captioning model - BLEU-4 (\textbf{B@4}) \cite{papineni2002bleu}, ROUGE-L (\textbf{R}) \cite{lin2004rouge}, METEOR (\textbf{M}) \cite{denkowski2014meteor} and CIDEr (\textbf{C}) \cite{vedantam2015cider}.

\subsection{Datasets} \label{exp:datasets}
\textbf{MSVD - } This dataset \cite{chen2011collecting} contains 1970 short video clips from Youtube annotated with multi-lingual captions. We use only the ones in English and, similar to prior work \cite{chen2018tvt}, we separate the dataset into 1,200 train, 100 validation and 670 test videos. \\


\noindent \textbf{ActivityNet Captions -} This dataset \cite{krishna2017dense} contains 20K videos amounting to 849 video hours with over 100K total descriptions. We follow the official train-validation splits, which assigns 10,009 and 4,917 videos respectively. We only use one of the two validation splits provided and use ground-truth proposals for each caption to evaluate our model. 

\subsection{Implementation Details}
\subsubsection{\textbf{Feature Processing}}To extract Inception-v4 and ResNet-152 features, we sample the videos at 25 fps. We use the networks pre-trained on ImageNet \cite{krizhevsky2017imagenet}, remove the last fully-connected layer and extract 1536-D and 2048-D feature vectors respectively. While extracting I3D features, pre-trained on the Kinetics dataset \cite{carreira2017quo}, we sample the videos at 25 fps and extract features for every set of 64 consecutive frames to obtain 1024-D feature vectors. Similarly, 2048-D features are extracted from R(2+1)D pre-trained on Kinetics \cite{hara3dcnns}, using 16 frame clips. We use a Faster-RCNN network \cite{ren2016faster, wu2019detectron2} with a ResNet-50 backbone pre-trained on the Visual-Genome dataset \cite{krishna2017visual} and extract object features of each frame after sampling at 5 fps to obtain 2048-D feature representations. To tokenize ground-truth captions we use WordPiece tokenization from BERT \cite{devlin2018bert} with a vocabulary consisting of nearly 30K tokens.

\subsubsection{\textbf{Model Settings }} The model dimension of the transformer is 512 with 8 heads each of size 64. The hidden size of the feedforward layer is set to 2048. We use 3 encoder layers for each encoder and 4 decoder layers. The model is optimized using Adam with a learning rate of $5e-5$ on MSVD and $5e-4$ on ActivityNet Captions. We set the number of memory slots, $m$ in the memory-augmented encoder to $40$ and train using a batch size of 128.



\subsection{Results} \label{exp:results}
In Table \ref{tab:comparing_features}, we observe the advantage of fused representations over the individual feature contributions. Since, each network captures different visual aspects such as spatial context from Inception-v4, spatio-temporal context from R(2+1)D and object-level semantics from Faster-RCNN, their weighted-additive fusion aggregates the important characteristics from each visual modality and enhances it. We empirically determine that fusing R(2+1)D and Inception-v4 features along with Faster-RCNN features yield better scores. 

 Our model achieves competent performance on both datasets as shown in Table \ref{tab:msvd} and Table \ref{tab:activitynet}. In Table \ref{tab:msvd}, we demonstrate the improvement in captioning performance gained by replacing a word-level tokenizer, that uses a vocabulary sorted by word-frequency, with the WordPiece tokenizer used in BERT \cite{devlin2018bert}. The gain in performance could probably be realized by the advantage of using subwords to counter unknown tokens and the long-tailed problem in a vocabulary distribution sorted by word-frequency. We further improve our captioning performance by employing self-critical sequence training (SCST) \cite{rennie2017self}- a variant of the REINFORCE algorithm on sequences sampled from beam search with a beam size of $5$. We use the CIDEr metric as rewards for the agent (WAFTM), and the mean of the rewards is used as a reward baseline to counter high variance gradients leading to more stable convergence.

%% file: tex/6.conclusion.tex
\section{Conclusion}
\label{sec:conclusion}

We present WAFTM - a transformer with memory-augmented encoders to contextualizes a priori knowledge and a weighted-additive fusion mechanism while decoding to obtain enhanced feature representations. After empirically establishing features that lead to better visual understanding,  we improve our model performance using WordPiece tokenization and SCST. Using this technique, we exhibit competent results on the MSVD and ActivityNet Captions datasets.